\documentclass[a4paper]{article}
\usepackage{iwslt18,amssymb,amsmath,epsfig}
\def\reg{{\rm\ooalign{\hfil
     \raise.07ex\hbox{\scriptsize R}\hfil\crcr\mathhexbox20D}}}

\usepackage{times}
\usepackage{latexsym}
\usepackage{makecell}
\usepackage{soul}
\usepackage{url}
\usepackage{color}
\usepackage[normalem]{ulem}

\newcommand\white[1]{{\color{white}#1}}

\title{Robust Neural Machine Translation for Clean and Noisy Speech Transcripts}

  \makeatletter
 \def\name#1{\gdef\@name{#1\\}}
 \makeatother
 \name{{\em 
Mattia Di Gangi$^{\star \dagger}$\hspace{0.3cm} 
 Robert Enyedi\hspace{0.3cm} 
 Alessandra Brusadin\hspace{0.3cm} 
 Marcello Federico \thanks{The first author carried out the work during an internship at Amazon. } }}
 
\address{$^{}$ Amazon AI, East Palo Alto - USA\\$^{\star}$Fondazione Bruno Kessler, Trento - Italy\\$^{\dagger}$ University of Trento, Italy }

\begin{document}
\maketitle

\begin{abstract}
  Neural machine translation models have shown to achieve high quality when trained  and fed with well structured and punctuated input texts. Unfortunately, the latter condition is not met in  spoken language translation, where the input is generated by an automatic speech recognition (ASR) system. In this paper, we study how to adapt a strong NMT system to make it robust to typical ASR errors. As in our application scenarios transcripts might be post-edited by human experts, we propose adaptation strategies to train a single system that can translate either clean or noisy input with no supervision on the input type. Our experimental results on a public speech translation data set show that adapting a model on a significant amount of parallel data including ASR transcripts is beneficial with test data of the same type, but produces a small degradation when translating clean text. Adapting on both clean and noisy variants of the same data leads to the best results on both input types.
\end{abstract}

\section{Introduction}
The recent quality improvements \cite{bojar2017findings,bojar2018findings} of neural machine translation (NMT) \cite{sutskever2014sequence,bahdanau2014neural} opened the way to commercial applications that can provide high-quality translations. The assumption is that the sentences to translate will be similar to the training data, usually characterized by properly-formed text in the two translation languages. Poor-quality sentences are usually considered ``noise'' and removed from the training set to improve the final quality \cite{junczys2018microsoft}. This practice is so common that a shared task has been devoted to it \cite{koehn2018findings}, which got attention from major industrial players in MT.  Thus, the major weakness of NMT lies in coping with noisy input, which is an important feature of a real-world application such as speech translation. 

The degradation of translation quality with noisy input has been widely reported in literature. Belinkov and Bisk \cite{belinkov2017synthetic} showed that the translation quality rapidly drops with both natural and synthetic noise. Karpukhin et al. \cite{ruiz2017assessing} have observed a correlation between recognition and translation quality in the context of speech translation. In both cases the degradation is mainly due to word errors, and following works have shown that inserting synthetic noise in the training data increases the robustness to the same kind of noise \cite{karpukhin2019training,sperber2017toward}.

\begin{table}[t]
    \centering
    \begin{tabular}{c}
         Most of the time, travellers worry about their luggage. \\
         Most of the time travellers worry about their luggage.
    \end{tabular}
    \caption{Example of sentence in which the meaning is changed by a punctuation mark.}
    \label{tab:example1}
\end{table}

In practice, ASR transcripts are not only noisy in the choice of words, but also come without punctuation.\footnote{ASR services generally include punctuation as an option.} Thus, in order to feed MT systems with ASR output there are two main options: 
\textit{i)} use a separate model that inserts punctuation (\textit{pre-processing}) \cite{peitz2011modeling}; 
or \textit{ii)} train the MT system on non-punctuated source data to resemble the test condition (\textit{implicit learning}). 
The first option is exposed to error propagation, as punctuation inserted in the wrong position can alter completely the meaning of a sentence (see example in Table~\ref{tab:example1}. 
The second option has shown to increase systems robustness \cite{sperber2017toward} when the input is provided without punctuation. The two approaches have shown to be equivalent in handling text lacking punctuation \cite{vandeghinstecomparison}, probably because they both rely only on plain monolingual text to recover 
the missing punctuation.

We consider here  application scenarios (e.g. subtitling) where the same NMT system has to operate under both clean and noisy input conditions, namely post-edited or raw ASR outputs. We start from the hypothesis that word and punctuation errors compound and should be addressed separately for an increased robustness to errors. To verify our hypothesis we use a strong NMT model and fine-tune it on a recently released speech corpus\footnote{Available at https://ict.fbk.eu/must-c/}~\cite{mustc19} of TED Talks. 

Our findings are the following: \textit{implicit learning} of punctuation helps to recover part of the errors in the source side, but training on ASR output is more beneficial. Training on clean and noisy data together leads to a system that can translate both clean and noisy data without supervision on the type of sentence and without degradation. 

\begin{table}[t]
\begin{tabular}{l|lll}
         & Train    & Validation & Test \\\hline
Words    &  5.7M    & 31.5K   &  54.5K  \\
Segments &  250K    & 1.3K    &  2.5K   \\
Audio    &  457h    & 2.5h    &  4.2h \\
\end{tabular}
\caption{The used English-Italian corpus of TED Talks.}
\label{tab:must-c}
\end{table}

\begin{table}[t]
\begin{tabular}{l|ccc}
                        & Clean & Noisy & Noisy-np \\\hline
\texttt{Gen}                    & 32.3 (30.7) & 24.5 (24.0)               & 20.6  (22.4) \\
\texttt{Ada} & 34.9 (32.9)  & 25.9 (25.6)               & 21.8 (24.6)                 
\end{tabular}
\caption{BLEU scores of large-data generic and adapted NMT systems with clean and noisy input. Scores in parentheses do not consider punctuation in both hypothesis and reference.}
\label{tab:initial_results}
\end{table}

\section{Robust NMT}
In this paper we are interested in building a single system to translate speech transcripts that can be either raw ASR outputs, or human post-editing of them.
We define our problem in terms of domain adaptation, and use the method known as \textit{fine-tuning} or \textit{continued training} \cite{luong2015stanford,farajian2016fbk,chu2017empirical}. A parent model is trained on large data from several domains and used to initialize \cite{thompson2018freezing} models for spoken language translation (TED talks) on two input conditions: \textit{clean}, and \textit{noisy}. The \textit{clean} domain is characterized by correct transcriptions of talks with proper punctuation, while the \textit{noisy} domain can contain machine-generated errors in words and punctuation. 
In-domain data can be given with or without punctuation (allowing \textit{implicit learning}). 
In a multi-domain setting, \textit{i.e.} translating both \textit{clean} and \textit{noisy} data with a single system, models can suffer from catastrophic forgetting \cite{kirkpatrick2017overcoming} by degrading performance on less-recently observed domains. In this work we avoid catastrophic forgetting by fine-tuning the model on both domains simultaneously.

\section{Experimental Setting}
We use as a parent model for fine-tuning a Transformer Big model \cite{vaswani2017attention} trained on public and proprietary data using label smoothed cross entropy \cite{szegedy2016rethinking}, for about 16 million English-Italian sentence pairs. The model has layer size of $1024$, hidden size of $4096$ on feed forward layers, 16 heads in the multi-head attention, and $6$ layers in both encoder and decoder. This model is then fine-tuned on the En$\rightarrow$It portion of TED Talks in MuST-C. We keep the same training/validation/test set split as provided with the corpus (see Table~\ref{tab:must-c}). In all the experiments, we use Adam \cite{kingma2014adam} with a fixed learning rate of $2\times10^{-4}$, dropout of $0.3$, label smoothing with a smoothing factor of $0.1$. Training is performed on $8$ Nvidia V100 GPUs, with batches of $2000$ tokens per GPU. Gradients are accumulated for $16$ batches in each GPU \cite{ott2018scaling}. All texts are tokenized and true-cased with scripts from the Moses toolkit \cite{koehn2007moses}, and then words are segmented with BPE \cite{sennrich2015neural} with 32K joint merge rules.

While our main goal is to verify our hypotheses on a large data condition, thus the need to include proprietary data, for the sake of reproducibility we also provide results 
with systems only trained on TED Talks (small data condition).

We transcribed automatically the entire TED Talks corpus with with a general purpose ASR service.~\footnote{Amazon Transcribe: https://aws.amazon.com/transcribe}
The resulting word error rates on the test set is 11.0\%. The used ASR service provides transcribed text with predicted punctuation. In the experiments which assume noisy input without punctuation, 
we simply remove the predicted punctuation.  

All the results are evaluated in terms of BLEU score \cite{papineni2002bleu} using the multeval tool \cite{clark2011better}.

\begin{table}[t]
\begin{tabular}{l|ll}
                          & Clean & Noisy  \\\hline
Clean                    & 34.9  & 25.9      \\
Clean-np                 & 34.2$^\dagger$  & 26.6$^\dagger$  \\
Clean + Clean-np         & 34.9  & 26.9$^\dagger$  \\\hline
Noisy                       & 34.0$^*$   & 28.3$^*$      \\
Noisy-np                    & 34.2$^*$   & 28.4$^*$      \\
Noisy + Clean              & 35.1$^\dagger$  & 28.1$^*$      \\
Noisy + Noisy-np           & 34.0$^*$   & 28.2$^*$      \\
Noisy-np + Clean         & 35.0$^\dagger$  & 28.2$^*$      \\
Noisy-np + Clean-np      & 34.5  & 27.9$^*$      \\
Noisy[-np] + Clean[-np] & 34.9$^\dagger$ & 27.7$^*$
\end{tabular}
\caption{Results of fine-tuning on different training conditions with clean and noisy input (large data). $^*$~means statistical significant (p $<$ 0.01) wrt to Clean + Clean-np, $^\dagger$~means statistical significant (p $<$ 0.01) wrt the first system of the block. (With randomization tests with 15K repetitions  \cite{riezler-maxwell-2005-pitfalls})}
\label{tab:fine-tuning}
\end{table}

\section{Experiments and Results}

At first, we evaluate the degradation due to ASR noise for systems trained on clean data. In Table~\ref{tab:initial_results} we show the BLEU scores of our baseline system, respectively, trained on large out-of-domain data (\texttt{Base}) and fine-tuned on clean TED Talks data (\texttt{In-domain}), with three types of input: manual transcripts (Clean), ASR transcripts with predicted punctuation (Noisy), and ASR transcripts with no punctuation (Noisy-np).
As these models will hardly generate punctuation when it does not appear in the source text, we also report (in parentheses) BLEU scores computed w/o punctuation in both hypothesis and reference. 

Our results show that in-domain fine-tuning is beneficial in all scenarios, but more with clean input (+2.6 points) than with noisy input (+1.4 points). Translating noisy input results in a 26\% relative drop in BLEU, which is apparently not due to punctuation errors (drop when evaluating w/o punctuation is 22\%).
Providing noisy input with no punctuation works even worse, probably due to the larger mismatch 
with the training/tuning conditions. 

\begin{table}[t]
\begin{tabular}{l|ll|ll}
                          & Clean & np & Noisy & np  \\\hline
Clean                    & 30.3 & - &  22.3  & -    \\
Clean-np                 & -  & 28.2 & - & 22.9    \\
Clean + Clean-np          & 29.7  & 27.9  &  22.9  & 22.9    \\\hline
Noisy                       & 25.8$^\dagger$  & - &  23.9$^\dagger$ & -     \\
Noisy-np                    & -  & 26.4  &  -  & 24.1$^\dagger$    \\
Noisy-np + Clean           & 30.1  & 27.9 & 24.0$^\dagger$ & 24.2$^\dagger$      \\
\end{tabular}
\caption{Results of fine-tuning on different training conditions with clean and noisy input (small data).$^\dagger$: statistical significant difference with Clean.}
\label{tab:fine-tuning-small}
\end{table}

\begin{table*}[th]
\small
\begin{tabular}{l|l}
\hline
Clean & when I 'm not fighting poverty , I 'm fighting fires as the assistant captain of a volunteer fire company . \\
Noisy & when I 'm not fighting poverty \uline{.} I 'm fighting fires \uline{.} \uline{is} the assistant captain \uline{with} volunteer fire company .\\
Base NMT & Quando non combatto la povert\`a  \uline{.} Combatto gli incendi \uline{. \`E} l'assistente capitano di una compagnia di pompieri volontaria .\\
Robust NMT & Quando non combatto la povert\`a , combatto gli incendi come assistente capitano di una compagnia di pompieri volontaria .  \\\\
\hline
Clean &  that means we all share a common ancestor , an evolutionary grandmother , who lived around six million years ago. \\
Noisy &  that means we all share a common ancestor \uline{\white{-}} \uline{on} evolutionary grandmother \uline{\white{-}} who lived around six million years ago. \\
Base NMT & Ci\`o significa che tutti condividiamo un antenato comune \uline{\white{-}} \uline{sulla} nonna evolutiva che ha vissuto circa sei milioni di anni fa. \\
Robust NMT & Ci\`o significa che tutti condividiamo un antenato comune \uline{e} una nonna evolutiva che \`e vissuta circa sei milioni di anni fa. \\\\\hline

Clean & and we in the West couldn't understand how anybody would do this , how much this would restrict freedom of speech. \\
Noisy & \uline{\white{---}} we in the West \uline{.} \uline{I} couldn 't understand how anybody would do \uline{\white{---}} \uline{\white{-}} how much this would restrict freedom of speech. \\
Base NMT & \uline{\white{---}} Noi occidentali \uline{.} Non riuscivo a capire come chiunque avrebbe fatto \uline{\white{---}} \uline{\white{-}} quanto questo avrebbe limitato la libert\`a di parola. \\
Robust NMT & \uline{\white{---}} In Occidente non riuscivo a capire come chiunque avrebbe fatto \uline{\white{---}} , quanto questo avrebbe limitato la libert\`a di parola. \\
\hline
\end{tabular}
\caption{Examples of punctuation and substitution errors ("as $\rightarrow$ is", "of a $\rightarrow$ with", "an $\rightarrow$ on") that are successfully recovered by the Robust NMT system. Notice that not all errors (underlined) are recovered. In the second example, Robust NMT introduces a spurious conjunction "e" ("and") in place of the missing comma, while in the third example, Robust NMT is not able to recover the deleted words "and" and "this" at the begin and in the middle of the sentence, respectively. }
\label{tab:example2l}
\end{table*}

Next, we evaluate systems fine-tuned on data similar to the Noisy test condition. Table~\ref{tab:fine-tuning} lists the results of all fine-tuning experiments, when test input 
is either clean with punctuation or noisy with punctuation as in training. 
The first part of Table~\ref{tab:fine-tuning} shows that fine-tuning the \texttt{Gen} model with clean data and no punctuation (Clean-np) improves over the \texttt{Ada} model (Clean) when testing on Noisy-np (+0.7) but degrades when testing on Clean (-0.7). Fine-tuning the same model on Clean data with both punctuation condition (Clean + Clean-np) improvement by 1 point on Noisy input without any loss on Clean input. This result shows that is possible to make a model robust to noisy text, while preserving high quality on proper text.

The second part of Table~\ref{tab:fine-tuning} lists results of fine-tuning on noisy data, with and/or without punctuation. In all cases, BLEU scores on the noisy input improve from 1 to 2 point over the best systems tuned on Clean data only, reaching values above 28. However, scores on clean input degrade by 0.7 and 0.9 points (i.e. 34.2 and 34.0). If we adapt on both clean and noisy data, the score on the two input conditions reach a better balance. In particular, training on both Noisy-np and Clean data scores on the two input conditions 35.0 and 28.2, which results in the best overall working point on both conditions (together with Noisy+Clean). 
It is worth pointing out that this configuration obtains 33.2 points (not in the table) with the best possible noisy input, \textit{i.e.} no errors and no punctuation, which is still 1.7 points below the score on Clean input.

Finally, if we expose the system to all types of data (Noisy-np~+Noisy+~Clean~+~Clean-np) we do not see any improvement over our top results, which means that Clean-np data do not provide additional information to Noisy-np.

For the sake of replicability, we also trained our systems from scratch on the TED Talks data only. The results, listed in Table~\ref{tab:fine-tuning-small}, show the same trend as the results discussed so far. We did not evaluate *-np systems on input with punctuation as all the punctuation would represent out of vocabulary words. The main difference resides in the result on Clean input with the Noisy system ($25.8$ points), which is much worse than the result with the Noisy-np+Clean system ($30.1$ points), i.e. more than 4 points. This result suggests how training on noisy data can affect the model negatively if it is not balanced with clean data.


\begin{table}[t]
\begin{tabular}{l|ll}
                       & ASR w ties & ASR w/o ties\\                       \hline
Clean                  & 10.5 & 32.8 \\
ASR-np + Clean         & 21.7$^\dagger$ & 67.2$^\dagger$  \\
\end{tabular}
\caption{Manual evaluation in the ASR input condition (large data). Percentage of wins with and without ties. $^\dagger$~stands for statistically significant (p $<$ 0.01).}
\label{tab:man-eval}
\end{table}

\section{Manual Evaluation}

We carried out a manual evaluation\footnote{We used crowd-sourcing via figure-eight.com.} 
to assess the quality of Noisy-np~+~Clean against Clean, the reference baseline, under the Noisy input condition. We ran a head-to-head evaluation on the first 10 sentences of each test talk, for a total of 260 sentences, by asking annotators to blindly rank the two system outputs (ties were also permitted). We collected three judgments for each output, from 11 annotators, for a total of 780 scores. Inter-annotator agreement measured with Fleiss' kappa was 0.39. 

Results reported in Table~\ref{tab:man-eval} confirm with high confidence the differences observed with BLEU: output of system Noisy-np+Clean is preferred 10\% time more often than output of system Clean, while almost 68\% of the time the two outputs are considered comparable.  

From some manual inspection, we found that translations by the robust system that are unanimously ranked best show that error recovery most likely occurs on punctuation and non-content words like articles, prepositions and conjunctions (see Tables \ref{tab:example2l}). In general, errors on content words that affect the meaning are not recovered.

\section{Related works}
A recent study \cite{chen2017mitigating} proposed to tackle ASR errors as a domain adaptation problem in the context of dialog systems. Domain adaptation for NMT has been widely studied in recent years \cite{chu2018survey}. 
In \cite{khayrallah-etal-2018-regularized}, fine-tuning was used to adapt NMT to multiple domains simultaneously, while in \cite{britz2017effective} adversarial adaptation is proposed for avoiding degradation in the original domain. 
Training on multiple domains simultaneously to prevent catastrophic forgetting is inspired by \cite{stojanov2019incremental}. They proposed an incremental learning scheme that trains the network with the data from previous tasks when a new task is learned, which we adapt to our multi-domain scenario.

Punctuation insertion based on monolingual text has attracted research works for a long time  \cite{huang2002maximum,matusov2006automatic,lu2010better,ueffing2013improved} and obtained recent improvements with deep networks \cite{cho2012segmentation,cho2015punctuation,tilk2015lstm,tilk2016bidirectional,salloum2017deep}. However, this approach, which is meant to make the output of ASR more readable for humans, cannot solve ambiguity due to missing punctuation. 

A more recent research line aims at using pauses and audio features to better predict punctuation \cite{christensen2001punctuation,klejch2017sequence,zelasko2018punctuation,nanchen2019empirical,yi2019self} although it has been shown that the use of pauses is highly dependent on the speaker \cite{igras2016structure}. In \cite{peitz2011modeling}, it is shown that \textit{implicit learning} of punctuation in MT systems is at least as good as inserting punctuation either in the input or output text, but they only studied the effect on correct input that has been deprived of punctuation, not on noisy input. On the other side, \cite{sperber2017toward} studied how to improve the robustness to misrecognized words, but did not study the effect of MT systems that are only robust to punctuation errors. We close the gap by studying the combined effect of misrecognized errors and missing punctuation, besides studying the robustness to noisy data affects the translation quality on clean input.  


\section{Conclusion}
We have studied the robustness to input errors of NMT systems for speech translation with a fine-tuning approach. We have observed that a system trained to learn implicitly the target punctuation can recover part of the quality degradation due to ASR errors up to 1 BLEU point. Fine-tuning on noisy input can instead improve by more than 2 BLEU points. A system tuned on ASR errors does not obtain a further improvement by more data for implicit punctuation learning. Finally, when fine-tuning on clean and noisy data, the system becomes robust to noisy input and keeps high performance on clean input.
\newpage
\bibliography{biblio}
\bibliographystyle{IEEEtran}

\end{document}